%% file: iclr2025_conference.tex
\documentclass{article} 
\usepackage{iclr2025_conference,times}

\input{math_commands.tex}

\usepackage{hyperref}
\usepackage{graphicx}
\usepackage{url}
\usepackage{ragged2e} 
\usepackage{tabularx} 
\usepackage{booktabs}
\usepackage{longtable}
\usepackage{tikz}
\usepackage{adjustbox}
\usepackage{multirow, multicol}

\title{FigmaTrace: Capturing Creative Nuances in Human Figma Design Workflows}

\author{Darshan Deshpande\thanks{Correspondence: \texttt{darshan@patronus.ai}}, Yoshinari Fujinuma, Martyna Markiewicz, Devanshu Bansal \\
\textbf{Shivani Jain}, \textbf{Nicholas Saban}, \textbf{Chirag Maheshwari}, \textbf{Anand Kannappan} \\
\includegraphics[width=0.3cm]{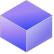} Patronus AI\\
\small{\texttt{\{darshan, yoshinari.fujinuma, martyna, dev, shivani,}} \\
\small{\texttt{nicksaban, chirag.maheshwari, anand\}@patronus.ai}}}

\newcommand{\datasetname}{\textsc{FigmaTrace}}

\iclrfinalcopy 
\begin{document}

\maketitle

\begin{abstract}
Vision Language Models have recently shown improvements in several objective and verifiable domains such as object detection but continue to underperform on subjective and creative design tasks. A major contributor to this performance gap is the lack of high quality human workflow data that captures a diverse set of preferences and decisions that make human experts good at design tasks. In this work, we first define a unique, expert curated taxonomy of design skills and best practices which we further expand into a set of 126 open ended, subjective, long horizon tasks. Built on top of this and expert solutions, our dataset~\datasetname~contains over 200 hours of human captured video data converted into 3469 design trajectories using a novel design phase-based method. We use our dataset to train four models and show that training on~\datasetname~leads to a performance improvement comparable to frontier closed models such as \textsc{Claude-Opus-5} and \textsc{GPT-5.6-Sol} on four out of distribution agentic GUI environments. We further perform a useful ablation to attribute these performance improvements to a design phase-based video to trajectory conversion which outperforms prior length-based conversion approaches. Finally, we perform a qualitative analysis on the best performing \textsc{Qwen3.8-27B} outputs to better correlate performance improvements to~\datasetname's trends. We open source our dataset and the best \textsc{Qwen3.8-27B} model for the community\footnote{\url{https://huggingface.co/datasets/PatronusAI/figmatrace} \\ \url{https://huggingface.co/PatronusAI/Qwen3.8-27B-Figmatrace-SFT}}. 
\end{abstract}

\section{Introduction}
Vision Language Models (VLMs) are popularly used for several verifiable tasks such as document understanding~\citep{ding-etal-2026-survey, wang2025documentintelligenceeralarge}, robotics~\citep{sapkota2025vision, zhang2025purevisionlanguageaction} and computer use~\citep{tang2025surveymllmbasedguiagents, xie2025large}. On subjective and non-verifiable tasks, VLMs have struggled to capture human nuances such as understanding emotions~\citep{bhattacharyya-wang-2025-evaluating}, humor and understanding of figurative meaning~\citep{zhou2026icameisaw, ryan-etal-2025-humor} and design taste~\citep{an2026visionlanguagemodelsassess}. Recent works have attempted to address this issue through human preference alignment~\citep{peng2025designprefcapturingpersonalpreferences, liao2025humanaesexpertadvancingmultimodalityfoundation} and reinforcement learning based objectives~\citep{li2025qinsightunderstandingimagequality, wu2026visualquality}, however, these techniques rely heavily on the availability of data that surfaces such preferences. This is worsened by the unavailability of high quality design datasets used to train \textit{tasteful} and \textit{nuanced} design agents.

\begin{figure*}[t]
  \centering
  \newcommand{\trajframe}[2]{%
    \begin{minipage}[t]{0.32\textwidth}
      \centering
      \includegraphics[width=\linewidth]{figmatrace_traj/#1}\\[1pt]
      {\footnotesize #2}
    \end{minipage}}
  \trajframe{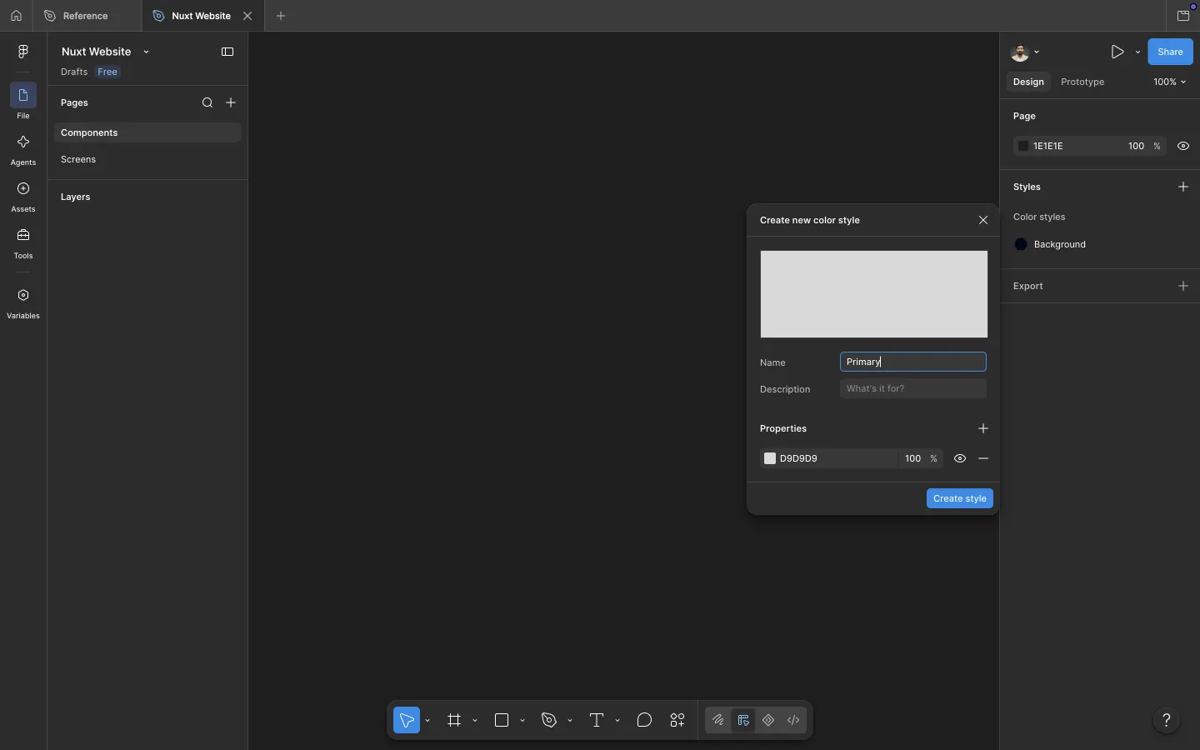}{(a) Set hexcode - \texttt{keyboard\_type("00DC82")}}\hfill
  \trajframe{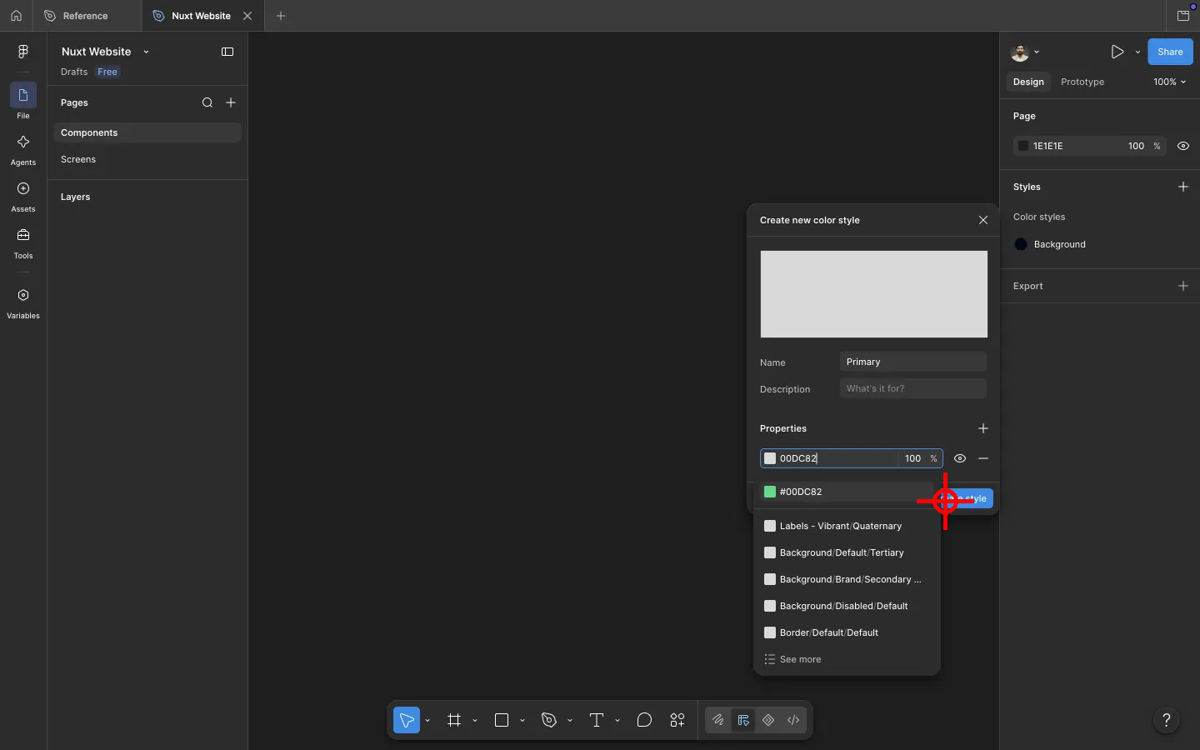}{(b) Create style - \texttt{mouse\_click(1134, 602)}}\hfill
  \trajframe{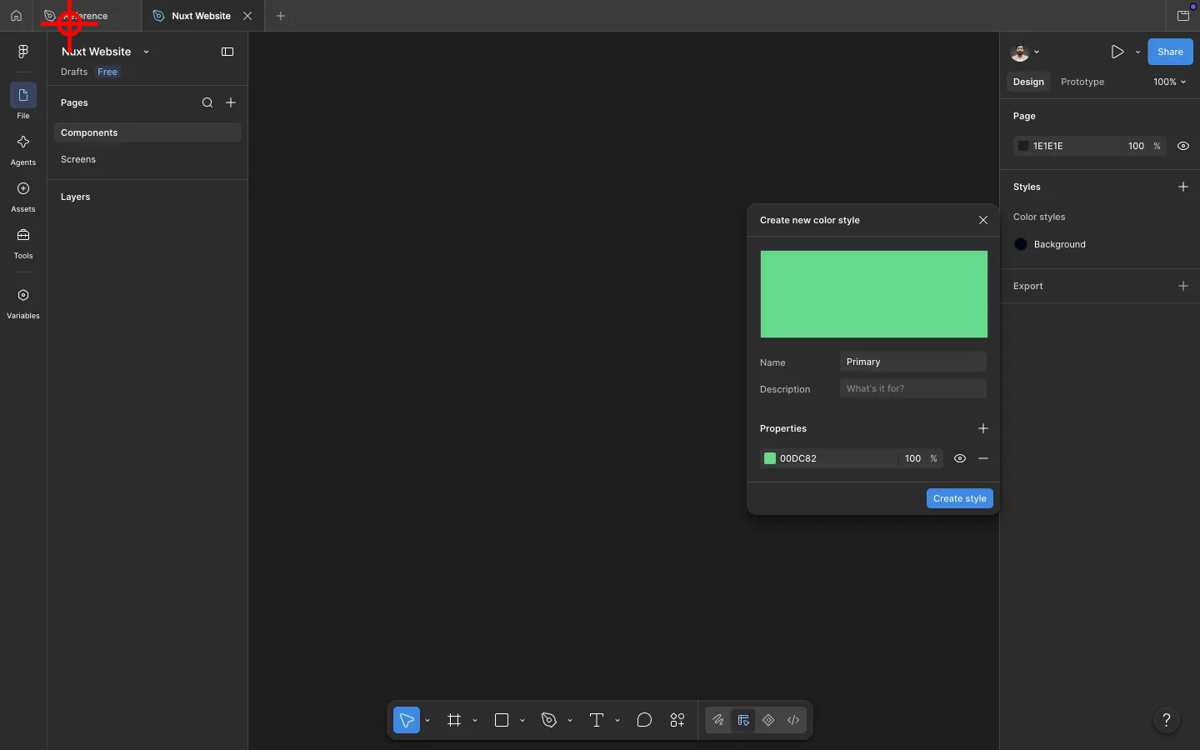}{(c) Switch tab - \texttt{mouse\_click(84, 28)}}\\[6pt]
  \trajframe{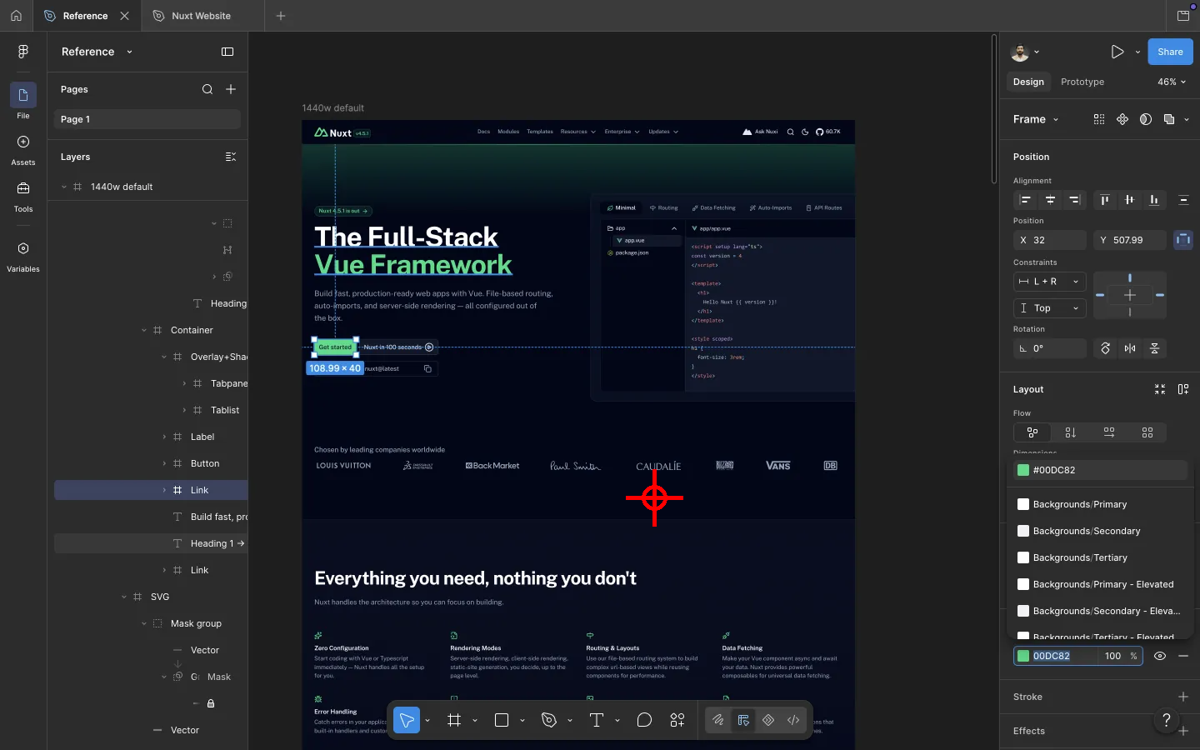}{(d) Select element - \texttt{mouse\_click(786, 598)}}\hfill
  \trajframe{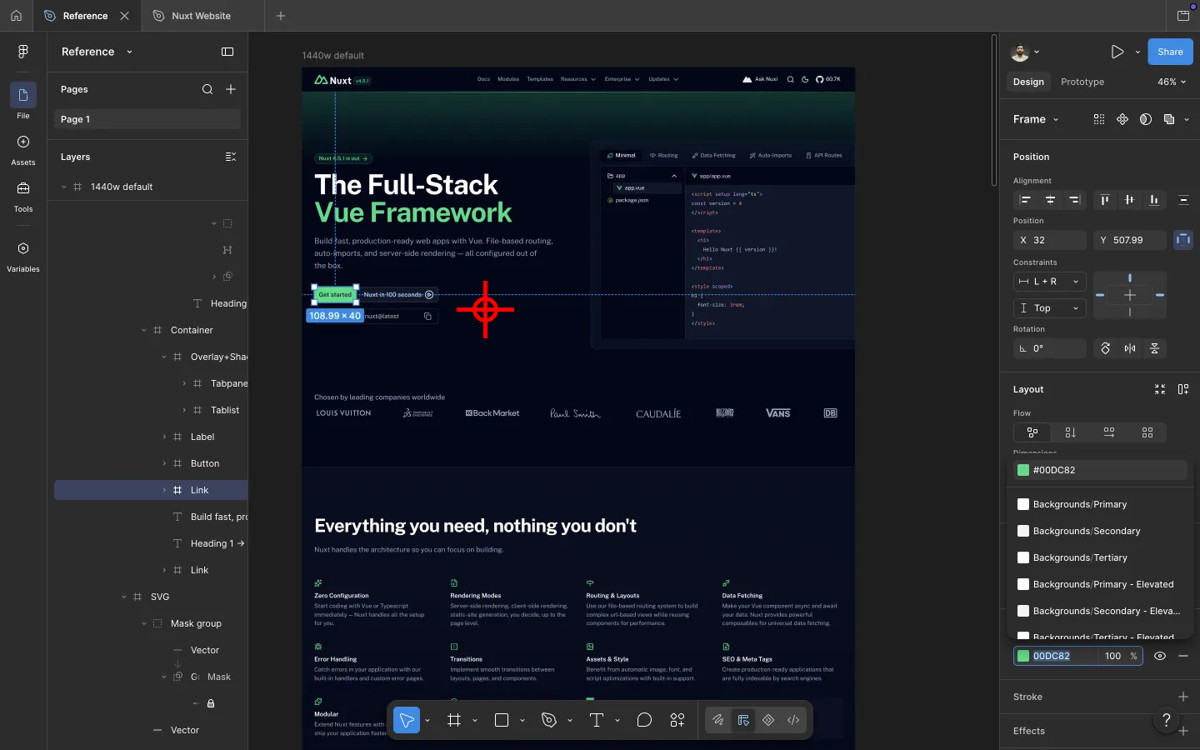}{(e) Select element - \texttt{mouse\_click(583, 372)}}\hfill
  \trajframe{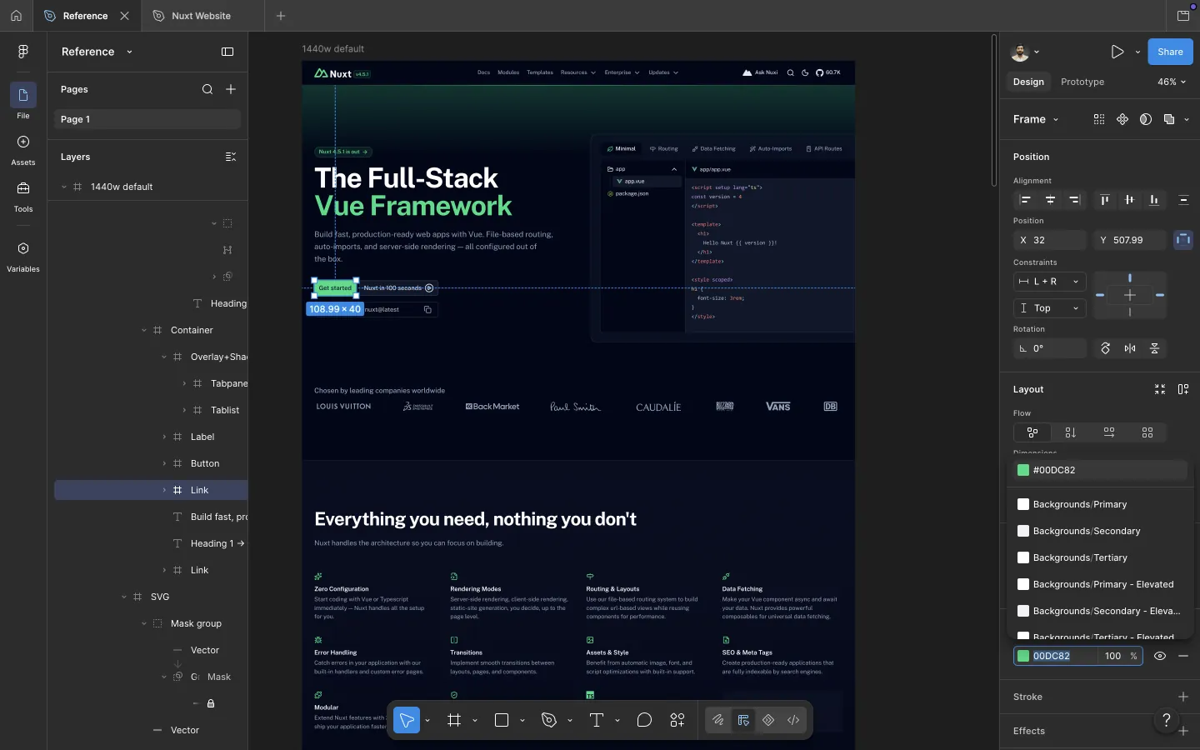}{(f) Resulting state}
  \caption{A sample~\datasetname~trajectory excerpt taken from session \texttt{t1-112},
  task involving replicating the Nuxt Website.}
  \label{fig:figmatrace-trajectory}
\end{figure*}

To address this lack of data, prior works such as~\citet{gui2026figmacode} explored using existing Figma designs and working backwards to create automated annotation processes for data but the validation of such work is difficult and ambiguous due to the lack of exhaustive quality guidelines. Exploring verifiability,~\citet{russo2025bridging} use existing HTML versions of pages to convert them to Figma compatible JSONs that a model can be trained to generate (human grounding for LLMs is difficult + validation is hard). ~\citet{kanapathipillai2026cogen} explore a parallel direction of automated creation of individual Figma component JSONs for the sake of reusability and scalability. Making the process of web design agentic,~\citet{jeong2026canvas} propose a harness for evaluating agents on design tasks but take no account of human taste involved in curating such tasks. Despite efforts on these fronts, evaluation of such data is increasingly difficult due to non-determinism of multimodal automated judges. ~\citet{chandwani2026lh} propose a unique set of skills that can guide the evaluation process but the scope of such pre-defined skills is limited and does not generalize effectively to design applications.  

To address these issues, we propose~\datasetname, a comprehensive dataset with 126 long horizon tasks, spanning across 8 realistic designer workflows, capturing a set of 10 high level expert curated skills split across 3469 trajectories. We first capture OS-level expert actions and screen captures, each working through unique task categories that cover both verifiable and open ended design tasks. Using this video dataset, we clean and thoroughly post process the recordings and corresponding actions to create a comprehensive set of ultra-long horizon workflow trajectories spanning up to 5 million tokens. To effectively train on these trajectories and retain the skills and design phases (such as creating components, reference gathering, etc) showcased in the trajectories, we create an expert-reviewer curated set of phase categories showcased in these workflows. We then use these categories to automatically extract and verify labels using \textsc{Gemini-3.6-Flash} at scale. Using our final dataset, we study the following research questions:

\begin{enumerate}
    \item Does training with realistic human captured design workflows teach VLMs to be better at agentic navigation and design?
    \item Does conversion of video data to trajectories benefit more from design-phase based trajectory curation as opposed to maximum context-length sharding for very long horizon tasks?
    \item What patterns in~\datasetname~influence qualitative performance improvements in models?
\end{enumerate}

Through our results, we show that training VLMs on~\datasetname~improves agent performance on several goal oriented and narrative oriented benchmarks including but not limited to GUI-Odyssey~\citep{lu2025guiodyssey}, Mind2Web~\citep{deng2023mind2web}, VideoGUI~\citep{lin2024videogui}, achieving an absolute increase of up to 46\% on datasets such as AndroidControl~\citep{li2024effects} over the corresponding baselines. Beyond this, we find that performance improvement brought about by our design phase-based trajectory curation approach leads to a 7.3\% absolute increase in performance, showing that the model understands task intents much better as compared to maximum context-length based SFT where the intent becomes unclear due to sharding at inconsistent intervals. Finally, we perform human evaluation to study the source of performance boost on out-of-domain tasks and find that the dataset improves properties such as element selection accuracy and decisiveness of model decisions in undirected settings.

\section{Related Work}

\paragraph{Design-to-code}
Rico~\citep{deka2017rico} is one of the earliest works on creating a design-focused mobile screens dataset with view hierarchies, followed by Design2Code~\citep{si-etal-2025-design2code} which curates real webpages for screenshot-to-HTML generation, WebSight~\citep{laurencon2024unlockingconversionwebscreenshots} which scales the same formulation to two million synthetically rendered pages, and \citet{gui2026figmacode} which extends it to Figma. In each case, the focus is on the final artifact rather than the trajectories for creating such artifacts, which lets an agent learn what a finished design looks like but not the sequence of decisions that produced it.

\begin{figure*}
    \centering
    \includegraphics[width=\linewidth]{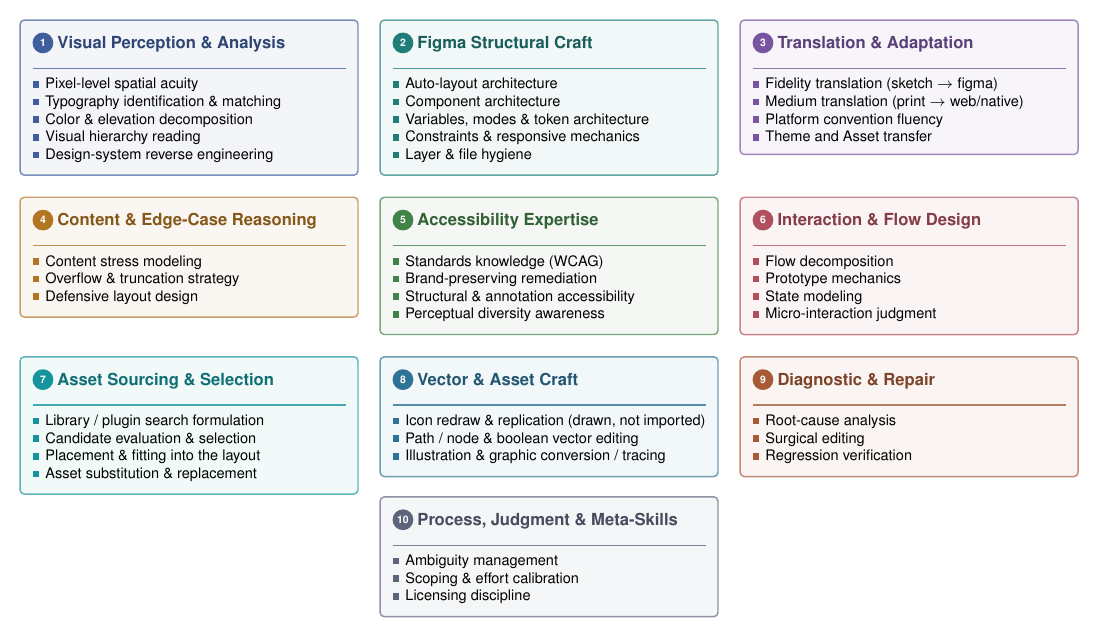}
    \caption{A comprehensive set of relevant skills that~\datasetname~covers.}
    \label{fig:sme_skills}
\end{figure*}

\paragraph{Screen Recordings for Computer Use Agents} Exploiting screen recordings has long been studied, but datasets including screen recordings and aligned gold actions are scarce. Where recordings are unavailable, grounding data is instead synthesized by decomposing and recomposing interfaces~\citep{xie2026scaling}, and evaluated on professional software by ScreenSpot-Pro~\citep{li2025screenspotpro}.
VideoAgentTrek~\citep{lu2025videoagenttrekcomputerusepretraining} uses public unlabeled screen recordings by first applying video-to-action mapping to detect actions on GUI to create synthetic agent trajectories.
OpenCUA~\citep{wang2025opencuaopenfoundationscomputeruse} instead records annotators directly, capturing screen video, input events, and the accessibility tree, and converts them into gold state-action pairs augmented with synthesized reasoning. 

\paragraph{Processing Screen Recordings}
Since raw screen recordings of human actions are often redundant and noisy, preprocessing is necessary to convert the recording into training data. 
VideoGUI~\citep{lin2024videogui} annotates instructional software video for evaluation at three levels: high-level planning, middle-level planning over action narrations, and atomic action execution.
VideoAgentTrek~\citep{lu2025videoagenttrekcomputerusepretraining} instead detects individual GUI actions with tight temporal bounds and attaches a per-action rationale.
Both are adequate when trajectories are short, but neither provides structure above the action, so intent boundaries become uncertain once a session runs for a longer horizon.

In summary, preprocessed screen recordings and actions showing the full expert trajectories are useful to train an agent from long-horizon sessions which require creative design skills. \datasetname~is the first dataset with pairs of gold action sequences and intent-segmented recordings of expert work for Figma.

\section{\datasetname}
This section describes the curation process of~\datasetname~ along with design decisions and expert feedback loops.

\newcolumntype{L}[1]{>{\RaggedRight\arraybackslash\hsize=#1\hsize}X}
 
\begin{table}
  \centering
  \small
  \caption{Taxonomy of Figma design task categories and their start-state
  creation methods.}
  \label{tab:task-taxonomy}
  \renewcommand{\arraystretch}{1.15}
  \begin{tabularx}{\linewidth}{@{} L{0.60} L{1.40} @{}}
    \toprule
    \textbf{ID / Task type} & \textbf{Start-state creation method} \\
    \midrule
    1.~Pixel-perfect replication (open seeds) &
    Pull a page from the open-web whitelist. Capture full-page PNG at 1440\,px
    (browser capture or SingleFile archive) and archive the URL, capture date,
    and license basis. Attach the one reusable instruction: ``Replicate 1:1 in
    Figma with proper auto-layout.'' \\
    \addlinespace
    2.~Responsive / platform adaptation &
    Attach a per-target instruction template: desktop $\rightarrow$ 375\,px
    mobile; web $\rightarrow$ tablet; print $\rightarrow$ web; web app
    $\rightarrow$ Android (Material). \\
    \addlinespace
    3.~Theming with variables &
    Instruction template on any cleared seed: ``Produce the dark/light variant
    implemented via Figma variables/modes---no manual per-node recolor.'' \\
    \addlinespace
    4.~Sketch-to-Figma &
    Start with a sourced hand-drawn sketch (existing hand-drawn-to-website
    datasets are also available). SME builds the hi-fi design from the photo.
    Both artifacts are owned via the contributor agreement. \\
    \addlinespace
    5.~Flaw injection $\rightarrow$ repair &
    Take an existing open-source Figma template and modify it to break a few
    things in the workflow. \\
    \addlinespace
    6.~Edge-content injection resilience &
    Test whether a layout survives content it was never designed for---most
    screens are only ever tested against clean, friendly demo data. \\
    \addlinespace
    7.~A11y remediation &
    Fix accessibility problems in the file. \\
    \addlinespace
    8.~Prototype wiring &
    From open-ended task generation, create a prototype of the website in
    Figma. \\
    \bottomrule
  \end{tabularx}
\end{table}

\subsection{Skill Taxonomy and Task Curation}
 
\paragraph{Taxonomy of Creative Skills}
To ground~\datasetname~ in real life creative workflows that human experts follow, we tasked three experts to create a comprehensive taxonomy of creative and nuanced human skills that designers utilize in their daily workflows.~\autoref{fig:sme_skills} showcases the taxonomy capturing 10 unique skills that cover best practices of Figma and are, in isolation or jointly, applicable to most Figma and non-Figma design applications. These cover core abilities of experts including visual perception, feature translation and adaptation, asset sourcing and creation, debugging, flow designing, accessibility best practices, content based reasoning and more. To the best of our knowledge, this is the most comprehensive taxonomy of design skills to date, thereby making~\datasetname~unique and useful for the community.

\paragraph{Task Coverage}
Inspired by realistic designer workflows, we design a set of eight unique task categories that strictly require one or more of the skills above. Specifically, these are categorized into verifiable and non-verifiable tasks~\autoref{tab:task-taxonomy}. Verifiable tasks include pixel perfect replication, repair of injected flaws, edge content resilience and a11y remediation that have deterministic solutions. On the other hand, tasks such as platform adaptation, theming, sketch to figma and prototype wiring capture nuance in workflows and hence outputs are dependent on SME biases, which in turn makes~\datasetname~a rich dataset.

\definecolor{barDark}{RGB}{23,54,93}    
\definecolor{barLight}{RGB}{208,224,240} 
\definecolor{barDarkTeal}{RGB}{23,93,93}
\definecolor{barLightTeal}{RGB}{208,240,240}
\definecolor{barDarkBurgundy}{RGB}{93,23,41}
\definecolor{barLightBurgundy}{RGB}{240,208,216}

\newcommand{\bpLabelW}{30}  
\newcommand{\bpNumW}{8}       
\newcommand{\bpGap}{7}       
\newcommand{\bpRowSep}{5.6}  
\newcommand{\bpBarH}{3}      

\newcommand{\bpfit}[1]{%
  \pgfmathsetmacro{\bpBarW}{\linewidth/1mm/#1 - \bpLabelW - \bpNumW - \bpGap}%
}

\newcommand{\barpanel}[5]{%
  \begin{tikzpicture}[
      x=1mm, y=-1mm,                            
      baseline=(current bounding box.north),  
      panelhead/.style = {font=\small\bfseries, anchor=west},
      panelnote/.style = {font=\footnotesize, text=black!55, anchor=east},
      rowlabel/.style  = {font=\footnotesize, anchor=east},
      rowvalue/.style  = {font=\footnotesize, anchor=west},
    ]
    \node[panelhead] at (-\bpLabelW,-2) {#1};
    \node[panelnote] at (\bpBarW+\bpNumW,-2) {#2};
    \draw[black!35,line width=0.3pt] (-\bpLabelW,1) -- (\bpBarW+\bpNumW,1);
    \foreach \share/\name [count=\row from 0] in {#5}{
      \pgfmathsetmacro{\y}{5 + \row*\bpRowSep}
      \pgfmathsetmacro{\len}{\share/#4*\bpBarW}
      \pgfmathsetmacro{\tone}{100 - 100*\row/(#3-1)}  
      \node[rowlabel] at (-1.5,\y) {\name};
      \fill[barDark!\tone!barLight]
        (0,\y-\bpBarH/2) rectangle (\len,\y+\bpBarH/2);
      \node[rowvalue] at (\len+1.5,\y) {\share};
    }
  \end{tikzpicture}
}

\newcommand{\barpanelteal}[5]{%
  \begin{tikzpicture}[
      x=1mm, y=-1mm,                            
      baseline=(current bounding box.north),  
      panelhead/.style = {font=\small\bfseries, anchor=west},
      panelnote/.style = {font=\footnotesize, text=black!55, anchor=east},
      rowlabel/.style  = {font=\footnotesize, anchor=east},
      rowvalue/.style  = {font=\footnotesize, anchor=west},
    ]
    \node[panelhead] at (-\bpLabelW,-2) {#1};
    \node[panelnote] at (\bpBarW+\bpNumW,-2) {#2};
    \draw[black!35,line width=0.3pt] (-\bpLabelW,1) -- (\bpBarW+\bpNumW,1);
    \foreach \share/\name [count=\row from 0] in {#5}{
      \pgfmathsetmacro{\y}{5 + \row*\bpRowSep}
      \pgfmathsetmacro{\len}{\share/#4*\bpBarW}
      \pgfmathsetmacro{\tone}{100 - 100*\row/(#3-1)}  
      \node[rowlabel] at (-1.5,\y) {\name};
      \fill[barDarkTeal!\tone!barLightTeal]
        (0,\y-\bpBarH/2) rectangle (\len,\y+\bpBarH/2);
      \node[rowvalue] at (\len+1.5,\y) {\share};
    }
  \end{tikzpicture}
}

\paragraph{Cleaning and Processing Action Spaces}
On average, we observed, through deterministic action to frame mapping that 95\% of actions captured during recordings were either random mouse movements or hover actions in the middle of the screen. Because our downstream agents use the Playwright MCP toolset,\footnote{\url{https://github.com/microsoft/playwright-mcp}} we filter out all mouse movements except hover actions. The toolset clicks a target directly from its screen coordinates, so no intermediate cursor movement is needed to reach it. We manually map all other OS level actions to the closest playwright-MCP action set. In some special cases such as when a render completes or when a plugin loads, screen state can change with no input. We insert an \verb|observe| probe every 2 seconds inside any gap of more than 4 seconds, so environment transitions become first-class steps.

\paragraph{Frame extraction}
In this step we extract frames from the video that correspond to the filtered actions above. We do this in two passes to ensure consistency: the first pass decodes the entire video (up to 4.5 hours long) at 8 frames per second in $480\times270$ grayscale. For each candidate at time $t$ this fixes two timestamps: before = $t - 0.15s$, and after = the first frame in $[t+0.2, t+2.0]$ where consecutive frames satisfy mean $|\Delta| < 0.75$ (the screen has settled). A fixed post-action offset is incorrect in this case since a menu settles in 0.25s on average (settle point found for 5,717/5,718 instances) and an image drop takes over a second. This pass acts as a proxy. The second pass extracts only those timestamps at full resolution, clustered into one decoder invocation per group, selecting exact frame indices so only the wanted frames are encoded. 

\paragraph{Effect Filtering}
For each pair, \texttt{changed\_fraction} = fraction of pixels whose max channel difference exceeds 6. Actions below a fraction of $5\times10^{-4}$ are dropped as having no visible effect whereas \verb|observe| probes need $2\times10^{-2}$ to count as a scene change. This is the step that separates what the expert did from what changed the artifact.

\paragraph{Skill based phase segmentation}
We utilize Gemini-3.6-Flash~\footnote{\url{https://blog.google/innovation-and-ai/models-and-research/gemini-models/gemini-3-6-flash-3-5-flash-lite-3-5-flash-cyber/}} as the video segmentation model. The model categorizes the video without the action log and returns contiguous spans from the closed 11 phase labels as described in~\autoref{tab:phase-taxonomy}. This incentivizes teaching skills to the VLM instead of randomly sharding based on pauses in the video that can be a noisy signal. The shard count that Gemini-3.6-Flash produces has no principled value, so rather than tune it we run 3, 6 and 12 shardings and keep only boundaries that $\ge 2$ of them place within $\pm30s$. This forms meaningful skill based separation which teaches VLMs specific skills required to learn Figma best practices. Finally, we assign skill labels to each trajectory based on frequency since one trajectory can potentially have more than one skill. During this segmentation process, we observed that video resolution matters far more than model used for segmentation. Against a strong model at full resolution (Gemini-3.6-Flash): same model at low resolution scores Jaccard 0.244 / 21\% dominant-skill agreement, while a weaker image model (Gemini-3-Pro as used by prior work) at full resolution scores 0.601 / 43\%. Low resolution collapses to generic labels because panel and layer text becomes unreadable. At the end of the entire process, we achieve a total compaction of $179\times$ as compared to the raw OS events captured by the screen recorder. 

\section{Experimental Setup}
\subsection{Training Setup}
To show performance improvements when training with~\datasetname, we use the ms-swift training framework~\citep{zhao2024swiftascalablelightweightinfrastructure} due to its strong support for long context training via context parallelization for VLMs. For showing consistent performance improvement we train \textsc{Qwen3.6-35BA3B}, \textsc{Qwen3.8-27B}, \textsc{Gemma-4-31B} and \textsc{Muse-Glimmer-30B} on 92,472 total actions sampled randomly from 35 sessions averaging at 47.6 hours of total human work done. A complete list of hyperparameters used can be found in the Appendix in~\autoref{tab:hyperparams}. To further study the generalization that~\datasetname~brings, we evaluate our fine-tuned models on four different dataset combinations, covering multi-app, multi-viewport GUI navigation using GUI-Odyssey~\citep{lu2025guiodyssey}, AndroidControl~\citep{li2024effects} to evaluate the effect of instruction granularity on model performance, Mind2Web~\citep{deng2023mind2web} to evaluate instruction grounding on open web data and VideoGUI~\citep{lin2024videogui}, a dataset testing planning and action narration and execution, sampled using video data that is not extracted from a skill based pattern. Since VideoGUI uses a different data post processing method to reshape videos into trainable data, improvement in performance on VideoGUI will also show the effectiveness and generalizability of our skill-based trajectory curation process. Since our trajectories do not come with pre-annotated reasoning chains, we train our models without reasoning. All closed model evaluations use high reasoning effort.  

  \begin{figure*}[t]
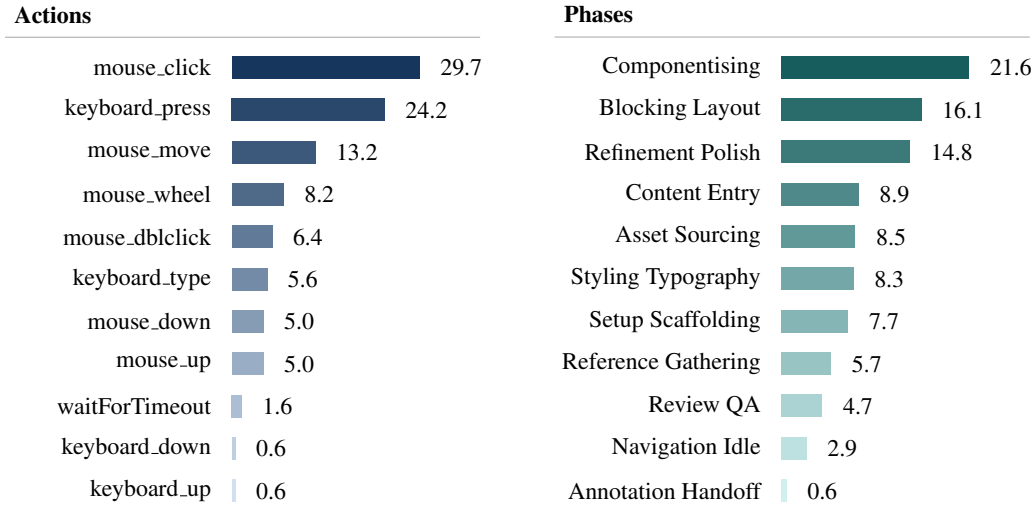

    \centering
    \bpfit{2}
    \makebox[\linewidth]{
      \barpanel{Actions}{}{11}{29.7}{
        29.7/mouse\_click,
        24.2/keyboard\_press,
        13.2/mouse\_move,
        8.2/mouse\_wheel,
        6.4/mouse\_dblclick,
        5.6/keyboard\_type,
        5.0/mouse\_down,
        5.0/mouse\_up,
        1.6/waitForTimeout,
        0.6/keyboard\_down,
        0.6/keyboard\_up}
      \hfill
      \barpanelteal{Phases}{}{11}{21.6}{
        21.6/Componentising,
        16.1/Blocking Layout,
        14.8/Refinement Polish,
        8.9/Content Entry,
        8.5/Asset Sourcing,
        8.3/Styling Typography,
        7.7/Setup Scaffolding,
        5.7/Reference Gathering,
        4.7/Review QA,
        2.9/Navigation Idle,
        0.6/Annotation Handoff}
    }
    \caption{\textbf{Combined-corpus composition in share percentage.} Actions follow the
    Playwright MCP toolset, counted over all validated tool calls. Phases come from the 11-way taxonomy as described in~\autoref{app:phase-taxonomy}. Bars are scaled within each panel and shares are rounded. The skill mix is listed out separately in \autoref{fig:skill-mix}.}
    \label{fig:composition}
  \end{figure*}

\section{Results and Discussion}

\paragraph{RQ1. Does training on~\datasetname~help improve next-action accuracy on design tasks?}~\autoref{tab:main-results} showcases the out-of-domain performance improvement seen for fine-tuned model as compared to the random, open and closed source model baselines. We observe that the~\datasetname~fine-tuned models compare favorably to state-of-the-art closed source models like \textsc{Claude-Opus-5} and \textsc{GPT-5.6-Sol} while being considerably smaller in size. Specifically, we observe that \textsc{Qwen3.8-27B} model outperforms even \textsc{Claude-Opus-5} at GUI-Odyssey and out of the box AndroidControl tasks by up to 6.4\% and 11.8\% absolute points. This shows that the navigation skills learned using~\datasetname~generalize to broader agentic and design tasks. As an additional in-domain design-task evaluation for the best performing \textsc{Qwen3.8-27B}, we utilize the ScreenSpot-Pro Creative~\citep{li2025screenspotpro} split and observed an absolute increase of performance of 7.4\% points over the base model performance (29.3\% vs 36.7\%).

\paragraph{RQ2. How does phase-based training of VLM agents compare against maximum length sharding for long horizon tasks?}

To investigate whether the phase-based trajectory creation process is beneficial, we compare this against the max length based truncation of trajectories. As observed in~\autoref{tab:phase-ablation}, we notice that phase-based outperforms maximum context length-based truncation by a margin of 7.3 absolute points. On closer qualitative analysis, we found that within AndroidControl the most affected rows are those that open mid-action, either as "Continue the work" stubs or as undirected references to on-page coordinates. This is because our phase-based method better teaches the model skill-based grounding instead of simple instruction following. For Mind2Web and VideoGUI, where the corpus focuses on layouts and not action grounding, the performance is not as significantly affected. Hence, our methodology for creating~\datasetname~outperforms maximum length-based trajectory generation processes overall. 

\begin{table}[t]
\centering
\small
\caption{Step-wise accuracy (\%) across OOD GUI agent benchmarks. For VideoGUI, directed refers to step-wise instruction provision whereas undirected is open-ended navigation. Mind2Web tests are run in two different image viewport (vp) configurations. \textbf{Bold}
indicates the best open model in each column and \underline{Underline} represents the best closed model. $^*$\textsc{Muse-Glimmer-30B} was evaluated similarly to the official evaluation with multiple zoom and crops averaged.}
\label{tab:main-results}
\begin{tabular}{@{}lrrrrrr@{}}
\toprule
 & & & \multicolumn{2}{c}{Mind2Web} & \multicolumn{2}{c}{VideoGUI} \\
\cmidrule(lr){4-5} \cmidrule(lr){6-7}
Model & \multicolumn{1}{c}{GUI-Odyssey} & \multicolumn{1}{c}{AndroidControl}
      & \multicolumn{1}{c}{vp800} & \multicolumn{1}{c}{vp1000}
      & \multicolumn{1}{c}{directed} & \multicolumn{1}{c}{undirected} \\
\midrule
Random Baseline  &  0.6           & 21.8           &  1.1           &  0.9           &  0.6           &  0.6 \\
\textsc{Claude-Opus-5}    & 47.3           & 87.3  & \underline{100.0}  & \underline{75.3}  & 71.3  & \underline{40.7} \\
\textsc{GPT-5.6-Sol}      & 44.0           & 83.2           & 91.3           & 69.7           & \underline{77.7}           & 29.6 \\
\midrule
\textsc{Qwen-3.6-35BA3B} & 29.0           & 36.8           & 38.2           & 30.7           & 47.0           & 10.1 \\
\textsc{Muse-Glimmer-30B} & 29.3 & 59.3 & 64.7 & 65.3 & 64.7 & 28.7 \\
\textsc{Gemma-4-31B}  & 43.3           & 87.3          & 68.0           & 64.0           & 63.0 & 24.0\\
\textsc{Qwen-3.8-27B}  & 44.0           & 82.7           & 69.3           & 65.3           & 65.3           & 26.0 \\
\midrule
\textsc{Qwen-3.6-35BA3B + SFT}             & 51.1  & 83.2           & 65.2           & 63.7           & 70.4           & 15.9 \\
\textsc{Muse-Glimmer-30B + SFT$^{*}$} & 53.2 &  84.5 & 70.2 & \textbf{70.1} & \textbf{73.0} & \textbf{32.8} \\
\textsc{Gemma-4-31B + SFT} & 45.2 & 96.1  & 69.1 & 66.0 & 67.0 & 23.3 \\
\textsc{Qwen-3.8-27B + SFT}             & \textbf{53.7}  & \textbf{99.1}           & \textbf{70.7}           & 68.7           & 71.3           & 19.3 \\
\bottomrule
\end{tabular}
\end{table}

\begin{table*}[t]
\centering
\caption{Comparison of phase-aware SFT against the base \textsc{Qwen3.8-27B} model and a maximum-length based SFT baseline across six GUI agent benchmarks. Best result per row in \textbf{bold}.}
\label{tab:phase-ablation}
\begin{tabular}{lccc}
\toprule
Benchmark & Base & Phase-based SFT & Length-based SFT \\
\midrule
GUI-Odyssey         & 44.0 & \textbf{53.7} & 40.0 \\
AndroidControl  & 82.7 & \textbf{99.1} & 67.3 \\
Mind2Web vp800      & 69.3 & \textbf{70.7} & 69.3 \\
Mind2Web vp1000     & 65.3 & 68.7          & \textbf{70.0} \\
VideoGUI-undirected       & \textbf{26.0} & 19.3          & 21.3 \\
VideoGUI-directed  & 65.3 & \textbf{71.3} & 70.0 \\
\midrule
\textbf{Mean}       & 58.8 & \textbf{63.8} & 56.3 \\
\bottomrule
\end{tabular}
\end{table*}

\paragraph{RQ3. What patterns in~\datasetname~influence qualitative performance improvements in mod-
els?} We manually inspect every item on which SFT converts a base failure into a success and find three recurring patterns. First, \textbf{element selection accuracy}: two-thirds of all gains observed on GUI-Odyssey are cases in which the base model selects an entirely different UI element. This is seen via the base model's median error, $\approx$ 457 px, while SFT lands within $\approx$ 15 px. The most illustrative case is the instruction "chat about it with a friend on Instagram", whose target is the message input at the bottom of the screen. For this setting, the base model predicts (596, 112), the search bar at the top of the frame, at essentially the same horizontal position as the gold point (593, 1325), whereas SFT lands within 7 px from the ground truth target. Such corrections concentrate on share icons, video cards, and chat inputs, which is reflected in the largest per-category improvements, including Media (+22 pp) and Social (+17 pp). Second, \textbf{coordinate understanding}: the base model frequently emits raw pixel coordinates instead of the norm-1000 coordinates it was originally post-trained on. A prediction of (800,212), for instance, falls directly on the share icon in pixel space but scores 360,px off once read as norm-1000. Especially in tall frames, targets in the bottom third overflow the grid entirely. In this case, raw value $y > 1000$ extends beyond the viewport frame. This occurs on 10/150 analyzed GUI-Odyssey items for the base model and on none for SFT. Third, \textbf{decisiveness}: every gain on AndroidControl comes from an item on which the base model either emits no coordinates at all or selects a totally incorrect element whereas SFT always answers, and when it corrects the element choice it lands at an average of $\approx$ 11 px.

On the other hand, where SFT fails, the same inspection applied to items on which SFT converts a base success into a failure reveals two newly acquired habits. The first is \textbf{repetition}: on Android app flows, SFT predicts effectively the same pixel on two consecutive steps, (331, 989) followed by (331, 988), while the ground truth trajectory advances down the list. The base model consistently tracks this progression correctly. We believe that this is an artifact of the noisy actions and mouse clicks that are leaked into~\datasetname~during preprocessing and conversion from raw video and action data. This behavior does not generally impact the overall task performance but results in a longer trajectory, thereby putting strain on the large context memory of the agent. The second is \textbf{screen-center focus}. Here, targets in the browser chrome are abandoned in favor of content in the middle of the screen, plausibly a leak of the canvas-centric bias induced by~\datasetname. Both of these categories cluster in utility and browser flows.

\section{Conclusion}
In this paper, we propose a novel skill taxonomy for general purpose design tasks and extend this to a novel dataset with 2883 training and 586 evaluation trajectories. Through our experiments, we show that training on~\datasetname~leads to performance improvement not only on the Figma design tasks but also on out of domain agentic tasks. Furthermore, we show that our phase-based video to trajectory conversion method outperforms the standard maximum context length extraction method. Finally, we perform qualitative analysis of model behaviors to categorize the success and failure modes that influence model performance on downstream tasks.

\bibliography{iclr2025_conference}
\bibliographystyle{iclr2025_conference}

\appendix
\section{Appendix}

\subsection{Phase Taxonomy}
\label{app:phase-taxonomy}

A \textbf{phase} is a contiguous stretch of a recording throughout which the expert holds a single intent. Phases tile the recording exactly and every frame belongs to exactly one phase. The label vocabulary is closed and free-form labels drift between calls. Table~\ref{tab:phase-taxonomy} gives the full vocabulary.

\subsection{Selection of Experts}
All subject matter experts (SMEs) contracted for this study were required to satisfy the following conditions: 1) Have a minimum of two years of experience with Figma, 2) Must be at least 18 years of age. Since all SMEs were hired through Upwork~\footnote{\url{https://upwork.com}}, we assigned every SME a starter task to vet the quality of their work. 

\subsection{Skill Distribution of the Dataset}
The distribution of skills is presented in~\autoref{fig:skill-mix}. We observe that structural craft dominates the distribution overall with visual perception being the second most frequent category. This is expected given that auto-layout and component hygiene are primary requirements for Figma design workflows.  

\newcommand{\barpanelBurgundy}[5]{%
  \begin{tikzpicture}[
      x=1mm, y=-1mm,                            
      baseline=(current bounding box.north),  
      panelhead/.style = {font=\small\bfseries, anchor=west},
      panelnote/.style = {font=\footnotesize, text=black!55, anchor=east},
      rowlabel/.style  = {font=\footnotesize, anchor=east},
      rowvalue/.style  = {font=\footnotesize, anchor=west},
    ]
    \node[panelhead] at (-\bpLabelW,-2) {#1};
    \node[panelnote] at (\bpBarW+\bpNumW,-2) {#2};
    \draw[black!35,line width=0.3pt] (-\bpLabelW,1) -- (\bpBarW+\bpNumW,1);
    \foreach \share/\name [count=\row from 0] in {#5}{
      \pgfmathsetmacro{\y}{5 + \row*\bpRowSep}
      \pgfmathsetmacro{\len}{\share/#4*\bpBarW}
      \pgfmathsetmacro{\tone}{100 - 100*\row/(#3-1)}  
      \node[rowlabel] at (-1.5,\y) {\name};
      \fill[barDarkBurgundy!\tone!barLightBurgundy]
        (0,\y-\bpBarH/2) rectangle (\len,\y+\bpBarH/2);
      \node[rowvalue] at (\len+1.5,\y) {\share};
    }
  \end{tikzpicture}
}
\begin{figure}[h]
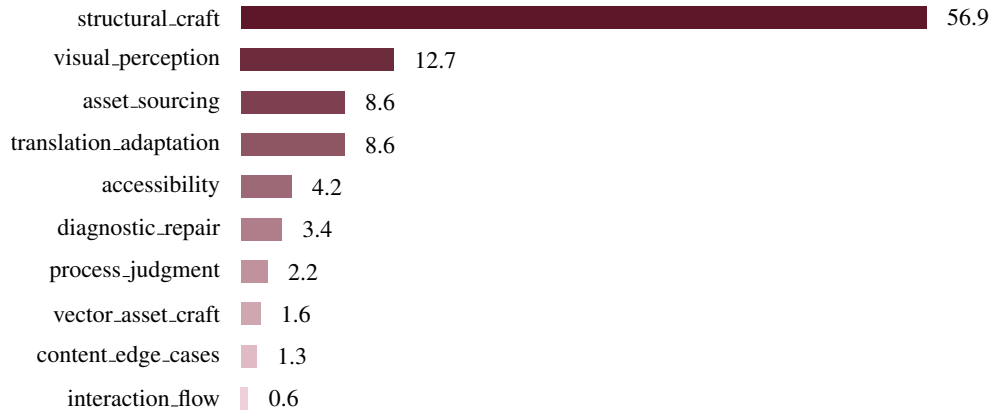

    \centering
    \renewcommand{\bpLabelW}{34}
    \bpfit{1}
    \barpanelBurgundy{Skills}{}{10}{56.9}{%
      56.9/structural\_craft,
      12.7/visual\_perception,
      8.6/asset\_sourcing,
      8.6/translation\_adaptation,
      4.2/accessibility,
      3.4/diagnostic\_repair,
      2.2/process\_judgment,
      1.6/vector\_asset\_craft,
      1.3/content\_edge\_cases,
      0.6/interaction\_flow}
    \caption{\textbf{Skill mix by share percentage} across all skill-labeled trajectories in~\datasetname.}
    \label{fig:skill-mix}
  \end{figure}

\begin{table}[t]
\centering
\caption{The closed phase vocabulary. Labels are ordered by their typical position in the workflow rather than by frequency.}
\label{tab:phase-taxonomy}
\small
\begin{tabularx}{\linewidth}{@{}lXr@{}}
\toprule
Label & Definition & Share (\%) \\
\midrule
\texttt{reference\_gathering} & Studying or collecting source material                      & 5.7  \\
\texttt{setup\_scaffolding}   & Establishing frames, artboards, grids, palettes, and styles & 7.7 \\
\texttt{blocking\_layout}     & Setting coarse structure, placement, and sizing             & 16.1 \\
\texttt{asset\_sourcing}      & Searching for or importing images, icons, and fonts        & 8.5  \\
\texttt{content\_entry}       & Typing real copy into the artifact                          & 8.9  \\
\texttt{styling\_typography}  & Making colour, type, and spacing decisions                  & 8.3  \\
\texttt{componentising}       & Turning ad hoc elements into reusable components            & 21.6 \\
\texttt{refinement\_polish}   & Making small deliberate adjustments and alignment passes    & 14.8 \\
\texttt{review\_qa}           & Comparing against reference; inspecting and checking        & 4.7  \\
\texttt{annotation\_handoff}  & Writing notes and comments; documenting decisions           & 0.6  \\
\texttt{navigation\_idle}     & Scrolling, waiting, or app churn with no creative intent    & 2.9  \\
\bottomrule
\end{tabularx}
\end{table}

\subsection{Reviewers and Annotators}
\input{creative_checklist}

\subsection{\datasetname~trajectory sample}
We present five consecutive recorded actions and the resulting state in~\autoref{fig:figmatrace-trajectory}. The expert types a hex value into the color-style dialog, commits the new style (\texttt{\#00DC82}), switches to the reference page, and selects elements to apply and verify it. Red crosshairs mark the recorded action coordinate on the frame the action was taken from. Every action-frame pair shown passed the corpus validation described in our curation method.

\subsection{Hyperparameters}

\autoref{tab:hyperparams} enumerates the best hyperparameters for all training runs reported in~\autoref{tab:main-results}. All runs reported in the paper were done with full-parameter SFT with the vision tower frozen and the aligner trainable. All runs are done with BF16 precision, Adam
($\beta{=}0.9/0.95$, weight decay $0.1$), gradient clipping $1.0$, cosine schedule with $3\%$ warmup, micro-batch $1$, and a $1{,}296{,}000$-pixel visual budget ($\approx$1{,}260 visual tokens/frame for \textsc{Qwen}, 1{,}120 soft tokens/image for \textsc{Gemma-4}). All training rows carry up to 44 frames. Megatron arms use the distributed optimizer with bf16 moment states and full activation recomputation; \textsc{Gemma-4} cannot use Megatron because its per-layer KV-head counts are heterogeneous, so it runs HuggingFace SFTTrainer with Deepspeed. The updated configuration for those runs includes per-device batch size of 1 and gradient accumulation of 2 for the Deepspeed ZeRO-3 runs.

\begin{table}[h]
  \centering
  \small
  \setlength{\tabcolsep}{4pt}
  \begin{adjustbox}{max width=\textwidth}
  \begin{tabular}{llllr}
    \toprule
    \textbf{Model} & \textbf{Arm} & \textbf{Backend} & \textbf{Parallelism} &
    \textbf{LR} \\
    \midrule
    \multirow{4}{*}{\textsc{Qwen3.6-35B-A3B}}
      & final (published) & Megatron & TP1\,CP8\,EP8\,DP2 & $2{\times}10^{-6}$  \\
      & baseline          & Megatron & TP1\,CP8\,EP8\,DP2 & $1{\times}10^{-5}$ \\
      & baseline (low lr) & Megatron & TP1\,CP8\,EP8\,DP2 & $2{\times}10^{-6}$  \\
      & ViT unfrozen      & Megatron & TP1\,CP8\,EP8\,DP2 & $1{\times}10^{-5}$ \\
    \midrule
    \multirow{2}{*}{\textsc{Qwen3.8-27B}}
      & SFT               & Megatron & TP4\,CP2\,DP2      & $2{\times}10^{-6}$  \\
      & max-length-shard abl. & Megatron & TP4\,CP2\,DP2      & $2{\times}10^{-6}$  \\
    \midrule
    \multirow{3}{*}{\textsc{Gemma-4-31B-it}}
      & plain             & HF\,+\,ZeRO-3 & DP16          & $2{\times}10^{-6}$  \\
      & plain (low lr)    & HF\,+\,ZeRO-2 & DP16          & $4{\times}10^{-7}$ \\
      & portrait-aug      & HF\,+\,ZeRO-3 & DP8           & $4{\times}10^{-7}$ \\
    \midrule
    \multirow{5}{*}{\textsc{Muse-Glimmer-30B}}
      & ATEM              & Megatron & TP4\,CP2\,DP2      & $4{\times}10^{-7}$ \\
      & ATEM (high lr)    & Megatron & TP4\,CP2\,DP2      & $2{\times}10^{-6}$ \\
      & JSON format       & Megatron & TP4\,CP2\,DP2      & $2{\times}10^{-6}$  \\
      & JSON (low lr)     & Megatron & TP4\,CP2\,DP2      & $4{\times}10^{-7}$  \\
      & \textbf{combined} & Megatron & TP4\,CP2\,DP2      & $4{\times}10^{-7}$ \\
    \bottomrule
  \end{tabular}
  \end{adjustbox}
  \caption{Training hyperparameters for every SFT arm.}
  \label{tab:hyperparams}
\end{table}

\end{document}

%% file: math_commands.tex
\usepackage{amsmath,amsfonts,bm}

\def\eqref#1{equation~\ref{#1}}

\def\1{\bm{1}}

\DeclareMathAlphabet{\mathsfit}{\encodingdefault}{\sfdefault}{m}{sl}
\SetMathAlphabet{\mathsfit}{bold}{\encodingdefault}{\sfdefault}{bx}{n}



%% file: creative_checklist.tex
\begingroup
\footnotesize
\renewcommand{\arraystretch}{1.15}
\setlength{\LTleft}{0pt}
\setlength{\LTright}{0pt}

\newcommand{\notassessed}{Pass / Fail / N/A}

\newcommand{\sectionrow}[1]{%
  \addlinespace[8pt]
  \multicolumn{3}{@{}l}{\textbf{#1}}\\*
  \addlinespace[3pt]}

\begin{longtable}{@{}
    >{\raggedright\arraybackslash}p{0.42\textwidth}
    >{\raggedright\arraybackslash}p{0.13\textwidth}
    >{\raggedright\arraybackslash}p{0.37\textwidth}
  @{}}

\caption{Design file readiness checklist.}
\label{tab:design-file-readiness-checklist}\\
\toprule
\textbf{Criterion} & \textbf{Status} & \textbf{Notes} \\
\midrule
\endfirsthead

\multicolumn{3}{@{}l}{\tablename~\thetable{} --- \emph{continued from previous page}}\\
\toprule
\textbf{Criterion} & \textbf{Status} & \textbf{Notes} \\
\midrule
\endhead

\midrule
\multicolumn{3}{r@{}}{\emph{Continued on next page}}\\
\endfoot

\bottomrule
\endlastfoot

\sectionrow{1.~File Structure \& Agent Navigation}
Cover page with file name, version, and status
  & Nice to have
  & Agent does not use the cover page directly, but it helps human reviewers orient quickly before handing the file over. \\
\addlinespace
Pages are logically organized (e.g. Cover / Components / Screens / Archive)
  & Pass/Fail & In accordance with Figma best practices\\
\addlinespace
Pages are not overloaded and large files are split across multiple pages to avoid exceeding agent context limit
  & Pass/Fail
  & Source: Figma MCP docs specifically mention `If you call \texttt{get\_design\_context} on an entire page instead of a specific node, the response can easily exceed the 25,000-token context window.` and this is expected to reduce this single call load. \\
\addlinespace
Each screen is a separate top-level frame with a descriptive name (e.g. "Login Screen", "Home / Mobile")
  & Pass/Fail
  & Source: Figma MCP documentation mentions agent uses \texttt{get\_metadata} to navigate the file by reading frame names. For example, "Frame 247" gives no orientation. \\
\addlinespace
Canvas contains no loose elements outside production frames i.e. no stray shapes, old iterations, or leftovers
  & Pass/Fail
  & Source: Figma MCP's \texttt{get\_metadata} scans the full page. Loose elements appear as noise and could disorient the agent. \\
\addlinespace
Old versions and iterations are archived on a separate page or removed. They should not be left on the working canvas.
  & Pass/Fail & This incentivizes cleanup actions for the agent trying to solve tasks. \\
\addlinespace
Pages and layers are named meaningfully. No default or autogenerated names
  & Pass/Fail & - \\
\addlinespace
No hidden elements or orphaned frames in the final submission file
  & Pass/Fail & - \\

\sectionrow{2.~Components \& Design System}
All components, styles, and assets live in the submission file. There should not be any cross-file library dependencies
  & Pass/Fail
  & For all of these tasks, the agent is expected to work within a single file. External libraries are not accessible. \\
\addlinespace
It should be clear which elements are components and which are layout frames. For example, repeated UI (buttons, inputs, cards, chips) must be real components but one-off layout frames (sections, containers, screen layouts) do not need to be.
  & Pass/Fail
  & Not everything needs to be a component. What matters is that repeating elements are components so the agent can reuse them, and one-off frames are intentionally not components and not accidentally detached. \\
\addlinespace
Accidentally detached instances are resolved i.e. if something was a component and got detached without intent, it should be reconnected or rebuilt.
  & Pass/Fail & - \\
\addlinespace
Components expose editable text properties (button labels, card copy) and text is not baked into the component.
  & Pass/Fail & - \\
\addlinespace
Component names are semantic and searchable. For example, "Button/Primary", "Card/Product" or "Input/Text"
  & Pass/Fail
  & Source: Figma MCP using agent uses \texttt{search\_design\_system} to find components by name. "Component 47" will not be found when searching for "button". \\
\addlinespace
Variant and property names are semantic. For example, "State=Default/Hover/Disabled" not "Property 1=Option 1/Option 2"
  & Pass/Fail
  & Source: Figma naming guide informs that each item is a gap the agent will fill with guesswork if you leave it. \\
\addlinespace
Auto Layout is applied to all relational components. No fixed/absolute positioning
  & Pass/Fail & This is for consistency purposes and in accordance with Figma best practices \\
\addlinespace
Constraints are set deliberately on elements that should respond to resizing
  & Pass/Fail
  & Not everyone, but it is not required in this case. \\
\addlinespace
File includes higher-order compositions (cards, headers, form rows) and not only atomic components (buttons, inputs)
  & \notassessed
  & Source: Figma Help Center mentions: "Atomic components are difficult for AI to compose into coherent layouts on their own." \\
\addlinespace
Component names use slash notation consistently accoridng to best practices. For example: Button/Primary/Large, Card/Product, Input/Text/Default
  & \notassessed
  & Source: Figma Help Center states "Figma slash notation creates nested groups that help both people and agents navigate large systems." \\
\addlinespace
Key components have a description filled in about what it is, when to use it, when not to use it, and keywords
  & Nice to have
  & Source: Figma MCP documentation: "Figma MCP reads component descriptions and passes them to the agent as context." Format: "[What it does]. Use for [when]. Do not use for [when not]. Keywords: [searchable terms]." Example: "Primary action button. Use for the main CTA on any screen. Do not use for secondary actions or destructive actions. Keywords: button, CTA, submit, confirm." This is a process decision. Refer to the next item. \\

\addlinespace[6pt]
\multicolumn{3}{@{}l}{\textit{Process recommendation --- component descriptions}}\\*
\addlinespace[3pt]
Who fills in component descriptions and when?
  & \notassessed
  & Component descriptions are not a natural part of the designer workflow because they are hidden in the component panel and rarely filled in without a specific process. Three options: (A)~designer fills in during component creation which requires process change and training, hard to maintain. (B)~one dedicated review before handing the file to the agent where someone goes through all components and adds descriptions; one-time effort, not ongoing. (C)~agent generates descriptions and agent reviews components and proposes descriptions that the designer approves; recommended for our setup. \\

\sectionrow{3.~Icons \& Assets}
Icons are embedded in the file as components and not linked from an external library the agent may not have access to
  & \notassessed & - \\
\addlinespace
Icons are not rasterized as images and not substituted with unicode characters or emoji
  & \notassessed & -\\
\addlinespace
Images use IMAGE fill type and not SOLID color standing in as a placeholder
  & \notassessed & -\\

\sectionrow{4.~Color-Reference to Design System}
Brand colors exist as color styles with correct values and nothing resolves to white or a wrong theme by default
  & \notassessed & -\\
\addlinespace
No hardcoded hex values on production frames. They should all be colors reference named styles
  & \notassessed & -\\
\addlinespace
Colors are also defined as Variables (tokens) for semantic referencing via MCP \texttt{get\_variable\_defs}
  & Nice to have
  & Source: Figma MCP docs:"\texttt{get\_variable\_defs} only returns tokens if the design uses them." More useful when agent works via MCP only. \\

\sectionrow{5.~Typography---Reference to Design System}
Shared text styles exist for the full hierarchy (heading / subheading / body / caption) and can be applied
  & \notassessed & -\\
\addlinespace
All fonts are available in Figma by default. No missing fonts (e.g. Proxima Nova is not embedded)
  & \notassessed & -\\
\addlinespace
No per-element font overrides. All text references named text styles
  & \notassessed & -\\

\sectionrow{6.~Spacing \& Layout}
The file has a discernible spacing scale which is ideally a documented token sheet (4 / 8 / 16 / 24px), at minimum a consistent scale evident in components
  & \notassessed & -\\
\addlinespace
Spacing and padding values are stored in Variables which enables semantic token referencing via MCP
  & Nice to have
  & Less critical if agent uses browser CUA where agent can read values visually. More useful when agent works via MCP only. \\

\sectionrow{7.~Interactive States}
Interactive components (buttons, inputs) include default, hover/active, and disabled variants
  & \notassessed & -\\
\addlinespace
Empty state is designed. Decisions are made about what the user sees when there is no content
  & \notassessed & -\\
\addlinespace
Error state is designed. Decisions are made about what happens when something goes wrong
  & \notassessed & -\\
\addlinespace
Loading state is designed. Decisions are made about what appears while content is loading asynchronously
  & \notassessed & -\\
\addlinespace
Components handle long text gracefully i.e. no overflow or broken layout at max content
  & \notassessed & -\\

\sectionrow{8.~Content \& Placeholder Quality}
Text content uses realistic placeholder copy and not "Lorem ipsum" or empty fields
  & \notassessed
  & Agent learns patterns from what it sees. Placeholder copy that resembles real content gives better context for tone, length, and information architecture. \\
\addlinespace
Images use realistic placeholder fills (IMAGE fill with a neutral image) and not empty rectangles or colored blocks
  & \notassessed & -\\
\addlinespace
Data in tables, lists, and cards represents realistic scenarios and not single-word entries or identical repeated rows
  & \notassessed & -\\

\sectionrow{9.~Accessibility}
Text contrast meets minimum ratio of 4.5:1 for body text, 3:1 for large text and interactive elements
  & \notassessed & -\\
\addlinespace
Focus states are designed for all interactive elements
  & \notassessed & -\\
\addlinespace
Text is not embedded in images. It must be readable as actual text layers
  & \notassessed & -\\

\sectionrow{10.~Annotations}
Each screen's purpose is clear from its content and naming alone. The agent should be able to understand what a screen does without reading a separate brief
  & \notassessed & -\\
\addlinespace
Non-obvious interactions are annotated: what triggers what, conditional logic, gestures
  & \notassessed & -\\
\addlinespace
Edge cases and constraints are noted where relevant (e.g. max character count, empty state triggers)
  & \notassessed & -\\

\sectionrow{11.~Final Completeness Check}
No missing icons. Every icon slot has an actual icon, not an empty frame or placeholder shape
  & \notassessed & -\\
\addlinespace
No missing images. Every image slot has an IMAGE fill, not an empty rectangle or colored block
  & \notassessed & -\\
\addlinespace
No missing copy. Every text layer has real or realistic placeholder content, not "Text", "Label", or empty strings
  & \notassessed & -\\
\addlinespace
No broken component instances. No instances showing "?" or missing component warnings in Figma
  & \notassessed & -\\
\addlinespace
No missing fonts. Figma shows no font warnings in the file
  & \notassessed & - \\
\addlinespace
All screens in scope are present. No screens referenced in flow but missing from the file
  & \notassessed & -\\
\addlinespace
All states for interactive components are present. Nothing is "TODO" or visually incomplete
  & \notassessed & -\\

\sectionrow{12.~Target-Size Fitness}
Components hold up visually at the file's target screen size (mobile 390px / desktop 1440px)
  & \notassessed & -\\
\addlinespace
Touch targets are at least 44$\times$44px for all interactive elements on mobile (according to best practices)
  & \notassessed & -\\

\end{longtable}
\endgroup

%% file: iclr2025_conference.bib
@inproceedings{deka2017rico,
    author = {Deka, Biplab and Huang, Zifeng and Franzen, Chad and Hibschman, Joshua and Afergan, Daniel and Li, Yang and Nichols, Jeffrey and Kumar, Ranjitha},
    title = {Rico: A Mobile App Dataset for Building Data-Driven Design Applications},
    year = {2017},
    isbn = {9781450349819},
    publisher = {Association for Computing Machinery},
    address = {New York, NY, USA},
    url = {https://doi.org/10.1145/3126594.3126651},
    doi = {10.1145/3126594.3126651},
    booktitle = {Proceedings of the 30th Annual ACM Symposium on User Interface Software and Technology},
    pages = {845–854},
    numpages = {10},
    location = {Qu{\'e}bec City, QC, Canada},
    series = {UIST '17}
}

@misc{laurencon2024unlockingconversionwebscreenshots,
      title={Unlocking the conversion of Web Screenshots into HTML Code with the WebSight Dataset}, 
      author={Hugo Laurençon and Léo Tronchon and Victor Sanh},
      year={2024},
      eprint={2403.09029},
      archivePrefix={arXiv},
      primaryClass={cs.HC},
      url={https://arxiv.org/abs/2403.09029}, 
}

@inproceedings{si-etal-2025-design2code,
    title = "{D}esign2{C}ode: Benchmarking Multimodal Code Generation for Automated Front-End Engineering",
    author = "Si, Chenglei  and
      Zhang, Yanzhe  and
      Li, Ryan  and
      Yang, Zhengyuan  and
      Liu, Ruibo  and
      Yang, Diyi",
    editor = "Chiruzzo, Luis  and
      Ritter, Alan  and
      Wang, Lu",
    booktitle = "Proceedings of the 2025 Conference of the Nations of the Americas Chapter of the Association for Computational Linguistics: Human Language Technologies (Volume 1: Long Papers)",
    month = apr,
    year = "2025",
    address = "Albuquerque, New Mexico",
    publisher = "Association for Computational Linguistics",
    url = "https://aclanthology.org/2025.naacl-long.199/",
    doi = "10.18653/v1/2025.naacl-long.199",
    pages = "3956--3974",
    ISBN = "979-8-89176-189-6"
}

@misc{wang2025opencuaopenfoundationscomputeruse,
      title={OpenCUA: Open Foundations for Computer-Use Agents}, 
      author={Xinyuan Wang and Bowen Wang and Dunjie Lu and Junlin Yang and Tianbao Xie and Junli Wang and Jiaqi Deng and Xiaole Guo and Yiheng Xu and Chen Henry Wu and Zhennan Shen and Zhuokai Li and Ryan Li and Xiaochuan Li and Junda Chen and Boyuan Zheng and Peihang Li and Fangyu Lei and Ruisheng Cao and Yeqiao Fu and Dongchan Shin and Martin Shin and Jiarui Hu and Yuyan Wang and Jixuan Chen and Yuxiao Ye and Danyang Zhang and Dikang Du and Hao Hu and Huarong Chen and Zaida Zhou and Haotian Yao and Ziwei Chen and Qizheng Gu and Yipu Wang and Heng Wang and Diyi Yang and Victor Zhong and Flood Sung and Y. Charles and Zhilin Yang and Tao Yu},
      year={2025},
      eprint={2508.09123},
      archivePrefix={arXiv},
      primaryClass={cs.AI},
      url={https://arxiv.org/abs/2508.09123}, 
}

@inproceedings{ding-etal-2026-survey,
    title = "A Survey on {MLLM}-based Visually Rich Document Understanding: Methods, Challenges, and Emerging Trends",
    author = "Ding, Yihao  and
      Luo, Siwen  and
      Dai, Yue  and
      Jiang, Yanbei  and
      Li, Zechuan  and
      Sun, Qiang  and
      Martin, Geoffrey  and
      Liu, Wei  and
      Peng, Yifan",
    editor = "Liakata, Maria  and
      Moreira, Viviane P.  and
      Zhang, Jiajun  and
      Jurgens, David",
    booktitle = "Findings of the {A}ssociation for {C}omputational {L}inguistics: {ACL} 2026",
    month = jul,
    year = "2026",
    address = "San Diego, California, United States",
    publisher = "Association for Computational Linguistics",
    url = "https://aclanthology.org/2026.findings-acl.652/",
    doi = "10.18653/v1/2026.findings-acl.652",
    pages = "13319--13340",
    ISBN = "979-8-89176-395-1"
}

@misc{wang2025documentintelligenceeralarge,
      title={Document Intelligence in the Era of Large Language Models: A Survey}, 
      author={Weishi Wang and Hengchang Hu and Zhijie Zhang and Zhaochen Li and Hongxin Shao and Daniel Dahlmeier},
      year={2025},
      eprint={2510.13366},
      archivePrefix={arXiv},
      primaryClass={cs.CL},
      url={https://arxiv.org/abs/2510.13366}, 
}

@misc{tang2025surveymllmbasedguiagents,
      title={A Survey on (M)LLM-Based GUI Agents}, 
      author={Fei Tang and Haolei Xu and Hang Zhang and Siqi Chen and Xingyu Wu and Yongliang Shen and Wenqi Zhang and Guiyang Hou and Zeqi Tan and Yuchen Yan and Kaitao Song and Jian Shao and Weiming Lu and Jun Xiao and Yueting Zhuang},
      year={2025},
      eprint={2504.13865},
      archivePrefix={arXiv},
      primaryClass={cs.HC},
      url={https://arxiv.org/abs/2504.13865}, 
}

@article{xie2025large,
  title={Large multimodal agents: a survey},
  author={Xie, Junlin and Chen, Zhihong and Zhang, Ruifei and Li, Guanbin},
  journal={Visual Intelligence},
  volume={3},
  number={1},
  pages={24},
  year={2025},
  publisher={Springer}
}

@article{sapkota2025vision,
  title={Vision-language-action models: Concepts, progress, applications and challenges},
  author={Sapkota, Ranjan and Cao, Yang and Roumeliotis, Konstantinos I and Karkee, Manoj},
  journal={arXiv preprint arXiv:2505.04769},
  year={2025}
}

@misc{zhang2025purevisionlanguageaction,
      title={Pure Vision Language Action (VLA) Models: A Comprehensive Survey}, 
      author={Dapeng Zhang and Jing Sun and Chenghui Hu and Xiaoyan Wu and Zhenlong Yuan and Rui Zhou and Fei Shen and Qingguo Zhou},
      year={2025},
      eprint={2509.19012},
      archivePrefix={arXiv},
      primaryClass={cs.RO},
      url={https://arxiv.org/abs/2509.19012}, 
}

@inproceedings{bhattacharyya-wang-2025-evaluating,
    title = "Evaluating Vision-Language Models for Emotion Recognition",
    author = "Bhattacharyya, Sree  and
      Wang, James Z.",
    editor = "Chiruzzo, Luis  and
      Ritter, Alan  and
      Wang, Lu",
    booktitle = "Findings of the Association for Computational Linguistics: NAACL 2025",
    month = apr,
    year = "2025",
    address = "Albuquerque, New Mexico",
    publisher = "Association for Computational Linguistics",
    url = "https://aclanthology.org/2025.findings-naacl.97/",
    doi = "10.18653/v1/2025.findings-naacl.97",
    pages = "1798--1820",
    ISBN = "979-8-89176-195-7"
}

@misc{zhou2026icameisaw,
      title={I Came, I Saw, I Explained: Benchmarking Multimodal LLMs on Figurative Meaning in Memes}, 
      author={Shijia Zhou and Saif M. Mohammad and Barbara Plank and Diego Frassinelli},
      year={2026},
      eprint={2603.23229},
      archivePrefix={arXiv},
      primaryClass={cs.CL},
      url={https://arxiv.org/abs/2603.23229}, 
}

@inproceedings{ryan-etal-2025-humor,
    title = "Humor in Pixels: Benchmarking Large Multimodal Models Understanding of Online Comics",
    author = "Ryan, Yuriel  and
      Tan, Rui Yang  and
      Choo, Kenny Tsu Wei  and
      Lee, Roy Ka-Wei",
    editor = "Christodoulopoulos, Christos  and
      Chakraborty, Tanmoy  and
      Rose, Carolyn  and
      Peng, Violet",
    booktitle = "Findings of the Association for Computational Linguistics: EMNLP 2025",
    month = nov,
    year = "2025",
    address = "Suzhou, China",
    publisher = "Association for Computational Linguistics",
    url = "https://aclanthology.org/2025.findings-emnlp.755/",
    doi = "10.18653/v1/2025.findings-emnlp.755",
    pages = "14024--14050",
    ISBN = "979-8-89176-335-7"
}

@misc{an2026visionlanguagemodelsassess,
      title={Can Vision Language Models Assess Graphic Design Aesthetics? A Benchmark, Evaluation, and Dataset Perspective}, 
      author={Arctanx An and Shizhao Sun and Danqing Huang and Mingxi Cheng and Yan Gao and Ji Li and Yu Qiao and Jiang Bian},
      year={2026},
      eprint={2603.01083},
      archivePrefix={arXiv},
      primaryClass={cs.CV},
      url={https://arxiv.org/abs/2603.01083}, 
}

@misc{peng2025designprefcapturingpersonalpreferences,
      title={DesignPref: Capturing Personal Preferences in Visual Design Generation}, 
      author={Yi-Hao Peng and Jeffrey P. Bigham and Jason Wu},
      year={2025},
      eprint={2511.20513},
      archivePrefix={arXiv},
      primaryClass={cs.CV},
      url={https://arxiv.org/abs/2511.20513}, 
}

@misc{liao2025humanaesexpertadvancingmultimodalityfoundation,
      title={HumanAesExpert: Advancing a Multi-Modality Foundation Model for Human Image Aesthetic Assessment}, 
      author={Zhichao Liao and Xiaokun Liu and Wenyu Qin and Qingyu Li and Qiulin Wang and Pengfei Wan and Di Zhang and Long Zeng and Pingfa Feng},
      year={2025},
      eprint={2503.23907},
      archivePrefix={arXiv},
      primaryClass={cs.CV},
      url={https://arxiv.org/abs/2503.23907}, 
}

@misc{li2025qinsightunderstandingimagequality,
      title={Q-Insight: Understanding Image Quality via Visual Reinforcement Learning}, 
      author={Weiqi Li and Xuanyu Zhang and Shijie Zhao and Yabin Zhang and Junlin Li and Li Zhang and Jian Zhang},
      year={2025},
      eprint={2503.22679},
      archivePrefix={arXiv},
      primaryClass={cs.CV},
      url={https://arxiv.org/abs/2503.22679}, 
}

@article{wu2026visualquality,
  title={Visualquality-r1: Reasoning-induced image quality assessment via reinforcement learning to rank},
  author={Wu, Tianhe and Zou, Jian and Liang, Jie and Zhang, Lei and Ma, Kede},
  journal={Advances in Neural Information Processing Systems},
  volume={38},
  pages={88167--88190},
  year={2026}
}

@inproceedings{
gui2026figmacode,
title={Figma2Code: Automating Multimodal Design to Code in the Wild},
author={Yi Gui and Jiawan Zhang and Yina Wang and Tianran Ma and Yao Wan and Shilin He and Dongping Chen and Zhou Zhao and Wenbin Jiang and Xuanhua Shi and Hai Jin and Philip S. Yu},
booktitle={The Fourteenth International Conference on Learning Representations},
year={2026},
url={https://openreview.net/forum?id=CaXZB6bI31}
}

@article{chandwani2026lh,
  title={LH-Bench: Skill-Grounded Evaluation of Long-Horizon Agents on Subjective Enterprise Tasks},
  author={Chandwani, Abhishek and Gupta, Ishan},
  journal={arXiv preprint arXiv:2603.22744},
  year={2026}
}

@inproceedings{russo2025bridging,
  title={Bridging Web and Figma: Automating Large-Scale UI Dataset Generation for AI-Enhanced Design},
  author={Russo, Francesca and Cal{\`o}, Tommaso and De Russis, Luigi},
  booktitle={Companion Proceedings of the 17th ACM SIGCHI Symposium on Engineering Interactive Computing Systems},
  pages={13--20},
  year={2025}
}

@article{kanapathipillai2026cogen,
  title={CoGen: Creation of Reusable UI Components in Figma via Textual Commands},
  author={Kanapathipillai, Ishani and Priyankara, Obhasha},
  journal={arXiv preprint arXiv:2601.10536},
  year={2026}
}

@inproceedings{jeong2026canvas,
  title={CANVAS: A Benchmark for Vision-Language Models on Tool-Based User Interface Design},
  author={Jeong, Daeheon and Byun, Seoyeon and Son, Kihoon and Kim, Dae Hyun and Kim, Juho},
  booktitle={Proceedings of the AAAI Conference on Artificial Intelligence},
  volume={40},
  number={26},
  pages={22182--22190},
  year={2026}
}

@misc{lu2025videoagenttrekcomputerusepretraining,
      title={VideoAgentTrek: Computer Use Pretraining from Unlabeled Videos}, 
      author={Dunjie Lu and Yiheng Xu and Junli Wang and Haoyuan Wu and Xinyuan Wang and Zekun Wang and Junlin Yang and Hongjin Su and Jixuan Chen and Junda Chen and Yuchen Mao and Jingren Zhou and Junyang Lin and Binyuan Hui and Tao Yu},
      year={2025},
      eprint={2510.19488},
      archivePrefix={arXiv},
      primaryClass={cs.CL},
      url={https://arxiv.org/abs/2510.19488}, 
}

@misc{zhao2024swiftascalablelightweightinfrastructure,
      title={SWIFT:A Scalable lightWeight Infrastructure for Fine-Tuning},
      author={Yuze Zhao and Jintao Huang and Jinghan Hu and Xingjun Wang and Yunlin Mao and Daoze Zhang and Zeyinzi Jiang and Zhikai Wu and Baole Ai and Ang Wang and Wenmeng Zhou and Yingda Chen},
      year={2024},
      eprint={2408.05517},
      archivePrefix={arXiv},
      primaryClass={cs.CL},
      url={https://arxiv.org/abs/2408.05517},
}

@inproceedings{
  li2025screenspotpro,
  title={ScreenSpot-Pro: {GUI} Grounding for Professional High-Resolution Computer Use},
  author={Kaixin Li and Meng Ziyang and Hongzhan Lin and Ziyang Luo and Yuchen Tian and Jing Ma and Zhiyong Huang and Tat-Seng Chua},
  booktitle={Workshop on Reasoning and Planning for Large Language Models},
  year={2025},
  url={https://openreview.net/forum?id=XaKNDIAHas}
}

@article{xie2026scaling,
  title={Scaling computer-use grounding via user interface decomposition and synthesis},
  author={Xie, Tianbao and Deng, Jiaqi and Li, Xiaochuan and Yang, Junlin and Wu, Haoyuan and Chen, Jixuan and Hu, Wenjing and Wang, Xinyuan and Xu, Yuhui and Wang, Zekun and others},
  journal={Advances in Neural Information Processing Systems},
  volume={38},
  year={2026}
}

@article{deng2023mind2web,
  title={Mind2web: Towards a generalist agent for the web},
  author={Deng, Xiang and Gu, Yu and Zheng, Boyuan and Chen, Shijie and Stevens, Sam and Wang, Boshi and Sun, Huan and Su, Yu},
  journal={Advances in Neural Information Processing Systems},
  volume={36},
  pages={28091--28114},
  year={2023}
}

@inproceedings{lin2024videogui,
  title={VideoGUI: A Benchmark for GUI Automation from Instructional Videos.},
  author={Lin, Kevin Qinghong and Li, Linjie and Gao, Difei and Wu, Qinchen and Yan, Mingyi and Yang, Zhengyuan and Wang, Lijuan and Shou, Mike Zheng},
  booktitle={NeurIPS},
  year={2024}
}

@article{li2024effects,
  title={On the effects of data scale on ui control agents},
  author={Li, Wei and Bishop, William and Li, Alice and Rawles, Chris and Campbell-Ajala, Folawiyo and Tyamagundlu, Divya and Riva, Oriana},
  journal={Advances in Neural Information Processing Systems},
  volume={37},
  pages={92130--92154},
  year={2024}
}

@inproceedings{lu2025guiodyssey,
  title={GUIOdyssey: A comprehensive dataset for cross-app GUI navigation on mobile devices},
  author={Lu, Quanfeng and Shao, Wenqi and Liu, Zitao and Du, Lingxiao and Meng, Fanqing and Li, Boxuan and Chen, Botong and Huang, Siyuan and Zhang, Kaipeng and Luo, Ping},
  booktitle={Proceedings of the IEEE/CVF International Conference on Computer Vision},
  pages={22404--22414},
  year={2025}
}
